\documentclass[runningheads]{llncs}
\usepackage{hyperref}
\usepackage{tikz}
\usepackage{booktabs}
\usepackage{subcaption}
\usepackage{algpseudocode}
\usepackage{algorithm}
\usepackage{stackengine}
\usepackage{amssymb}
\usepackage{multirow}
\usepackage{caption}
\usepackage{amsmath}
\usepackage{bm}
\usepackage{scalerel} 
\usetikzlibrary{shapes.geometric}
\usetikzlibrary{backgrounds}

\DeclareCaptionLabelFormat{andtable}{#1~#2  \&  \tablename~\thetable}

\algrenewcommand{\alglinenumber}[1]{\tiny#1:}

\begin{document}
\title{Searching the Space of Feed-Forward Neural-Network Weight-Update Rules with Fixed Depth Symbolic Regression}
\titlerunning{FFN Weight-Update Rules via Symbolic Regression}
%
\author{Edward Finkelstein\inst{1}
\and
Charlie Brum\inst{2}}
\authorrunning{Edward F. and Charlie B.} 
%
\institute{UCI, USA\\
\email{edfink234@gmail.com} \and San Diego, California, USA\\
\email{charlieb1658@gmail.com}}
\maketitle              
\begin{abstract}
We investigate whether symbolic regression can discover explicit neural network weight-update rules that outperform standard hand-designed optimizers on small symbolic regression benchmarks. Candidate update rules are represented as fixed-depth symbolic expressions over operands derived from common optimizers, including gradient, momentum, adaptive-gradient, and moment-estimate quantities. Across 30 benchmark/neural network combinations, the symbolic regression procedure found an update rule outperforming the best hyperparameter-tuned established optimizer in 25 cases, with an aggregate MSE reduction of 44.47\% over the improved cases. The discovered rules do not all share a single common symbolic form, but many combine adaptive normalization, momentum-like quantities, nonlinear transformations, and rational expressions. These results suggest that symbolic regression can serve as a lightweight mechanism for discovering compact optimizer variants, while also highlighting the need for larger-scale validation.

\end{abstract}

\section{Neural Network Weight-Update Rules}\label{sec:WeightUpdateRules}

Training a feed-forward neural network requires repeatedly updating its weights to reduce an error function. The basic gradient-descent update rule is:
\begin{equation}
    w_{t+1} = w_{t} - \alpha \nabla f(w_{t}),
    \label{eq:basic_update_rule}
\end{equation}
where $w_t$ denotes the weights at iteration $t$, $\alpha>0$ is the learning rate, and
\begin{equation}
    f = \frac{1}{N}||x(w)-y||^2 .
\end{equation}
Here, $x(w)$ is the prediction vector, $y$ is the target vector, and $N$ is the vector size.

Several widely used optimizers modify \eqref{eq:basic_update_rule} by incorporating momentum, adaptive learning rates, or moment estimates. Heavy-ball momentum is given by
\begin{align}
    v_{t+1} &= \theta v_{t} + \alpha \nabla f(w_{t}) \nonumber \\
    w_{t+1} &= w_{t} - v_{t+1},
    \label{eq:heavy_ball}
\end{align}
where $\theta \in [0,1]$ controls the influence of the previous velocity. Nesterov accelerated gradient instead evaluates the gradient at a look-ahead point:
\begin{align}
    v_{t+1} &= \theta v_{t} + \alpha \nabla f(w_{t}-\theta v_{t}) \nonumber \\
    w_{t+1} &= w_{t} - v_{t+1}.
\end{align}

Adaptive-gradient methods rescale updates using information from past gradients. AdaGrad is given by
\begin{equation}
    w_{t+1}=w_t - \alpha \,  g_{\Sigma} \circ g_t,
    \label{eq:Adagrad_rule}
\end{equation}
where $g_t = \nabla f(w_t)$, $\circ$ denotes the Hadamard product, and
\begin{equation}
    g_{\Sigma}
    =
    \left\{
    \left(\epsilon + \sum_{i = 1}^{t} g_{i, 1}^2\right)^{-1/2},
    \ldots,
    \left(\epsilon + \sum_{i = 1}^{t} g_{i, N}^2\right)^{-1/2}
    \right\}.
\end{equation}
RMSProp replaces the cumulative squared-gradient sum with an exponentially weighted running average:
\begin{align}
    w_{t+1} &= w_t + \Delta w_t \label{eq:RMSProp_update_rule} \\ 
    \Delta w_t &= - \frac{\alpha}{\sqrt{E[g^2]_t + \epsilon}} g_t \\
    E[g^2]_t &= \gamma E[g^2]_{t-1} + (1-\gamma) g_t^2 .
\end{align}

Adadelta applies a similar running-average normalization to both gradients and updates:
\begin{align}
    E\left[g^2\right]_t &= \gamma E\left[g^2\right]_{t-1} + (1-\gamma) g_t^2 \\
    \Delta w_t &= -\sqrt{\frac{E\left[\Delta w^2\right]_{t-1} + \epsilon}{E\left[g^2\right]_t + \epsilon}} g_t \\
    E\left[ \Delta w^2\right]_t &= \gamma E\left[\Delta w^2\right]_{t-1} + (1-\gamma) \Delta w_t^2 \\
    w_{t+1} &= w_t + \Delta w_t .
\end{align}

Adam combines a running average of past gradients with a running average of past squared gradients to adapt the update direction and scale:
\begin{align}
    \mu_t &= \beta_1 \mu_{t-1} + (1-\beta_1) g_t \\
    \nu_t &= \beta_2 \nu_{t-1} + (1-\beta_2) g_t^2 \\
    \widehat{\mu}_t &= \frac{\mu_t}{1-\beta_1^t} \\ 
    \widehat{\nu}_t &= \frac{\nu_t}{1-\beta_2^t} \\ 
    w_{t+1} &= w_t - \frac{\alpha \widehat{\mu}_t}{\sqrt{\widehat{\nu}^{\phantom{o}}_t+\epsilon}} .
    \label{eq:Adam_weight_update}
\end{align}
Finally, AdamW modifies Adam by applying regularization directly to the weights while leaving the adaptive moment estimates unchanged:
\begin{equation}
    w_{t+1} = w_t - \eta \lambda w_t - \frac{\alpha \widehat{\mu}_t}{\sqrt{\widehat{\nu}^{\phantom{o}}_t+\epsilon}} .
\end{equation}

The operands appearing in these optimizers motivate the symbolic regression search space used in this work.

\section{Applying Symbolic Regression to the Weight-Update Problem}\label{sec:Symbolic_Background}
Given that many of the weight update rules developed and currently used in practice are \emph{heuristically} motivated, an interesting question arises, namely, if algorithmically, one can find an empirically \emph{better} weight-update algorithm without significantly altering complexity and/or computational cost of existing ones. This section outlines the procedure of this paper to probe an answer to this question.
\subsection{Symbolic Regression}
Symbolic regression entails algorithmically searching a pre-specified basis-set of mathematical operators and operands for expressions which minimize a user-defined loss-metric, $\mathcal{L}$. In this paper, $\mathcal{L}$ is simply taken to be the mean-squared error between the labels $\vec{Y}$ of a dataset defined by the tuple $(\vec{X}, \vec{Y})$ and the output of a feed-forward neural network $f(\vec{X}, \vec{w})$ whose weights $\vec{w}$ shall be optimized. 
\par In this paper, the set of operators considered shall be the same for all tests conducted, given as follows:
\begin{itemize}
    \item \text{Unary Operators: } \textbf{-}, $\mathbf{ln}$, $\mathbf{exp}$, $\mathbf{cos}$, $\mathbf{sin}$, $\mathbf{sqrt}$, $\mathbf{asin}$, $\mathbf{acos}$, $\mathbf{tanh}$
    \item \text{Binary Operators: } \textbf{+}, \textbf{-}, \textbf{*}, \textbf{/}, $\bm{\wedge}$
    \item \text{Hyper-parameters: } $\bm{\eta}$, $\bm{\theta}$, $\bm{\epsilon}$, $\bm{\gamma}$, $\bm{\beta_1}$, $\bm{\beta_2}$
\end{itemize}

\subsection{Weight-Update Problem}
In section \ref{sec:WeightUpdateRules}, various weight-update rules were described, all of which can be represented by expression-trees using a pre-defined set of unary+binary operators and operands. Table \ref{tab:weight_update_rules_expression_trees} exemplifies this concept. 

\begin{table} 
    \centering
    \scalebox{0.8}{
    \begin{tabular}{|l|l|l|l|}
\hline 
Weight-Update Rule & Expression + Tree & Depth $N$ & \# of Inputs  \\ \hline 
  Gradient Descent &  \href{https://cb1658.github.io/Website/?img=5}{$w_{j,m,t=\tau} = w_{j,m,t=\tau-1} + \eta \cdot g_{j,m,t=\tau}$} & 2 & 3 \\[0.3cm]
  Heavy Ball  &  \href{https://cb1658.github.io/Website/?img=6}{$w_{j,m,t=\tau} = w_{j,m,t=\tau-1} + \theta \cdot v_{j,m,t=\tau-1} + \eta \cdot g_{j,m,t=\tau}$} & 3 & 5 \\[0.3cm]
  Nesterov  &  \href{https://cb1658.github.io/Website/?img=7}{$w_{j,m,t=\tau} = w_{j,m,t=\tau-1} + \theta \cdot v_{j,m,t=\tau-1} + \eta \cdot d_{j}^{\text{Nesterov}} \cdot y_{i,m,t=\tau}$} & 4 & 6 \\[0.3cm]
  AdaGrad  &  \href{https://cb1658.github.io/Website/?img=2}{$w_{j,m,t=\tau} = w_{j,m,t=\tau-1} + \dfrac{\eta \cdot g_{j,m,t=\tau}}{\sqrt{\sigma_{\hspace{-.05cm}g^{2}_{j,m}} + \epsilon}}$} & 4 & 5 \\[0.3cm]
   RMSProp  &  \href{https://cb1658.github.io/Website/?img=8}{$w_{j,m,t=\tau} = w_{j,m,t=\tau-1} + \frac{\eta \cdot g_{j,m,t=\tau}}{\sqrt{E\left[g_{j,m}^2\right]_{t=\tau} + \epsilon}}$} & 4 & 5 \\[0.3cm]
  AdaDelta  &  \href{https://cb1658.github.io/Website/?img=1}{$w_{j,m,t=\tau} = w_{j,m,t=\tau-1} - \Delta w^{\text{A}\hspace{-.018cm}\text{dadelta}}_{j,m,t=\tau}$} & 1 & 2 \\[0.3cm]
  Adam  &  \href{https://cb1658.github.io/Website/?img=3}{$w_{j,m,t=\tau} = w_{j,m,t=\tau-1} + \frac{\eta \cdot \widehat{\mu}_{j,m,t=\tau}}{\sqrt{\widehat{\nu}_{j,m,t=\tau}} + \epsilon}$} & 4 & 5 \\[0.3cm]
  AdamW  &  \href{https://cb1658.github.io/Website/?img=4}{$w_{j,m,t=\tau} = w_{j,m,t=\tau-1} + \eta \cdot \left(\lambda \cdot w_{j,m,t=\tau-1} + \frac{\widehat{\mu}_{j,m,t=\tau}}{\sqrt{\widehat{\nu}_{j,m,t=\tau}} + \epsilon}\right)$} & 6 & 6 \\[0.3cm] \hline
\end{tabular}}
    \caption[]{Weight update rules from section \ref{sec:WeightUpdateRules} and their corresponding minimal\footnotemark \phantom{ }expression tree information. Inputs in this table refer to unique leaf nodes in the tree. The expression trees are constructed based on the input operands defined for this paper in Table \ref{tab:operands}. An applet to scroll through the expression trees can be found \href{https://cb1658.github.io/Website/}{here}.} 
    \label{tab:weight_update_rules_expression_trees} 
\end{table}
\footnotetext{meaning the simplest (least number of nodes possible) expression tree one can construct to represent the algebraic formula}

The choice of operands, given in Table \ref{tab:operands}, reflects those utilized in the weight-update rules of section \ref{sec:WeightUpdateRules}, with the additional simplification that the studies conducted in this paper will only consider fully-connected feed-forward neural networks with either sigmoid ($f(x) = 1/(1+\exp(-x))$) and/or linear ($f(x) = x$) perceptron activation functions. 
\begin{table}
    \centering
    \begin{tabular}{|l|c|}
    \hline 
    \textbf{Operands}     &  \textbf{Description/Formula} \\[0.2cm] \hline
        $\tau$ & The current time-step ($t$) starting from 1. \\[0.15cm] 
       $w_{j,m,t=\tau}$  & The $m$'th weight of a neuron $j$ at time-step $t=\tau$ \\[0.15cm] 
       $d_{j}$ & The error term of a neuron $j$, i.e., $\partial f (w_{j, k}) / \partial x_j$ \\[0.15cm] 
       $y_j$ & Output of neuron $j$ \\[0.15cm] 
       $d_{j}^{\text{Nesterov}}$ & $\partial f(w_{j, k} - \theta v_{j, k}) / \partial x_j$  \\[0.15cm] 
       $v_{j, m, t=\tau}$ & The $m$'th velocity of a neuron $j$ at time-step $t=\tau$ \\[0.15cm] 
        $g_{j, m, t=\tau}$ & The $m$'th element of neuron $j$'s gradient vector at time-step $t=\tau$ \\[0.15cm] 
       $\sigma_{\hspace{-.05cm}g^{\scaleobj{0.75}{2}}_{\scaleobj{0.8}{j,m}}}$ & $\sum_{t = 1}^{\tau} g_{j, m, t}^2$, where $\tau$ is the current time-step/epoch\\[0.15cm] 
       $E\left[g_{j, m}^2\right]_{t=\tau}$ & $\gamma \cdot \left(E\left[g_{j, m}^2\right]_{t=\tau-1}\right) + (1-\gamma) g_{j, m, t=\tau}^2$, where $E\left[g_{j, m}^2\right]_{t = 0} \gets 0$ \\[0.15cm] 
       $\Delta w_{j, m, t=\tau}$ & $-\dfrac{\eta}{\sqrt{E\left[g_{j, m}^2\right]_{t=\tau} + \epsilon}} g_{j, m, t=\tau}$ \\[0.15cm] 
       $E\left[\Delta w_{j, m}^2\right]_{t=\tau}$ & $\gamma \cdot \left(E\left[\Delta w_{j, m}^2\right]_{t=\tau-1}\right) + (1-\gamma) \Delta w_{j, m, t=\tau}^2$, where $E\left[\Delta w_{j, m}^2\right]_{t = 0} \gets 0$ \\[0.15cm] 
       $\Delta w^{\text{A}\hspace{-.018cm}\text{dadelta}}_{j, m, t=\tau}$ & $-\sqrt{\dfrac{E\left[\Delta w_{j, m}^2\right]_{t=\tau} + \epsilon}{E\left[g_{j, m}^2\right]_{t=\tau} + \epsilon}} g_{j, m, t=\tau}$ \\[0.4cm] 
       $\mu_{j,m,t=\tau}$ & $\beta_1\cdot\mu_{j,m,t=\tau-1} + \left(1-\beta_1\right)\cdot g_{j, m, t=\tau}$, where $\mu_{j,m,t=0} = 0$ \\[0.15cm] 
        $\nu_{j,m,t=\tau}$ & $\beta_2\cdot\nu_{j,m,t=\tau-1} + \left(1-\beta_2\right)\cdot g^2_{j, m, t=\tau}$, where $\nu_{j,m,t=0} = 0$ \\[0.15cm] 
        $\widehat{\mu}_{j,m,t=\tau}$ & $\dfrac{\mu_{j,m,t=\tau}}{1-\beta_1^{t=\tau}}$ \\[0.35cm] 
        $\widehat{\nu}_{j,m,t=\tau}$ & $\dfrac{\nu_{j,m,t=\tau}}{1-\beta_2^{t=\tau}}$ \\[0.35cm]  
        $\eta$ & Learning rate hyperparameter \\[0.35cm]
        $\theta$ & hyperparameter for velocity scaling \\[0.35cm]
        $\gamma$ & hyperparameter for updating expectation value estimates \\[0.35cm] 
        $\epsilon$ & hyperparameter for stabilizing square-root terms \\[0.35cm]
        $\beta_1$ & hyperparameter for first-moment estimate update \\[0.35cm]
        $\beta_2$ & hyperparameter for second-moment estimate update \\[0.35cm]
        \hline 
    \end{tabular}
    \vspace{0.25cm}
    \caption{Input operands considered for the symbolic searches conducted in this paper. The symbol $f$ denotes the error function, $x_j = \sum_{i} y_i w_{j,i}$ is the non-activated input to neuron $j$, where $y_i$ are the individual outputs of the neurons $i$ from the previous layer, $w_{j,k}$ are the weights connecting neurons $j$ in the ``current'' layer to the neurons $k$ in the next layer. The hyperparameters are initialized arbitrarily as $\{\eta \gets 0.5, \, \theta \gets 0.01, \, \gamma \gets 0.9, \, \epsilon \gets 10^{-8}, \, \beta_1 \gets 0.9, \, \beta_2 \gets 0.999\}$; the symbolic evolution in practice automatically forms combined constants anyway.}
    \label{tab:operands}
\end{table}

\subsection{Symbolic Regression Method}\label{subsec:symb_reg_method}
The symbolic regression method utilized in this paper is the multi-threaded implementation of the generalized fixed-depth postfix grammar and the accompanying Genetic Programming algorithm utilized in the previous works \cite{finkelstein2024solving2dadvectiondiffusionequation} and \cite{finkelstein2024generalizedfixeddepthprefixpostfix}. The accompanying repository for this paper, \href{https://github.com/edfink234/Alpha-Zero-Symbolic-Regression/tree/NeuralNetworkWeightUpdate}{here}, includes a driver code that generates symbolic expressions and, for each weight-update rule expression, evaluates its ``score'' by updating the weights of a randomly initialized feed-forward neural network for an integer number of epochs using the generated weight-update rule, and computing its loss thereafter. The codes utilized for the experiments conducted in this paper are hyperlinked and described below:
\begin{itemize}
    \item \href{https://github.com/edfink234/Alpha-Zero-Symbolic-Regression/blob/cfbb9cb7f10ec1746a7ec1c33de9abb9a368f35a/NeuralNetworks_VecSR.cpp}{\texttt{NeuralNetworks\_VecSR.cpp}}: The file that implements the genetic-programming evolution and generation of candidate weight-update rules. 
    \item \href{https://github.com/edfink234/Alpha-Zero-Symbolic-Regression/blob/cfbb9cb7f10ec1746a7ec1c33de9abb9a368f35a/MLP_Vec.h}{\texttt{MLP\_Vec.h}} and \href{https://github.com/edfink234/Alpha-Zero-Symbolic-Regression/blob/cfbb9cb7f10ec1746a7ec1c33de9abb9a368f35a/MLP_Vec.cpp}{\texttt{MLP\_Vec.cpp}}: The files that implement the multi-layer perceptron class and weight-update rule logic.
    \item \href{https://github.com/edfink234/Alpha-Zero-Symbolic-Regression/blob/cfbb9cb7f10ec1746a7ec1c33de9abb9a368f35a/RunTestsNeuralNetworksVecSRPart1.py}{\texttt{RunTestsNeuralNetworksVecSRPart1.py}}: Driver code for conducting the hyperparameter grid-sweep for the established weight-update rules we benchmark against; see Table \ref{tab:hyper_param_grid_searches_done}.
    \item \href{https://github.com/edfink234/Alpha-Zero-Symbolic-Regression/blob/cfbb9cb7f10ec1746a7ec1c33de9abb9a368f35a/RunTestsNeuralNetworksVecSRPart2.py}{\texttt{RunTestsNeuralNetworksVecSRPart2.py}}: Driver code for executing the 30 different symbolic genetic evolutions for each of the 10 benchmarks and 3 feed-forward neural network architectures considered herein.
\end{itemize}

The primary contribution of this work is therefore an empirical study showing that symbolic regression with a grammar that produces expressions of a fixed expression-tree depth \cite{finkelstein2024generalizedfixeddepthprefixpostfix} can discover compact, explicit weight-update rules that outperform hyperparameter-tuned standard optimizers on a collection of small neural network symbolic regression tasks.

\section{Background}\label{sec:Background}

Discovering the optimal parameters of a feed-forward neural network entails devising a method to update them. To this end, many novel ideas have emerged over the years. For example, \cite{gregor2020finding} employs neural networks to learn weight and activation function update rules and shows the optimality of such rules to depend on problem complexity and neuron type(s). In \cite{andrychowicz2016learninglearngradientdescent}, the authors employ Long-short term memory (LSTM) neural networks as surrogates to the traditional gradient descent weight-update rule and show that this approach can generalize to similar examples on which the LSTM-updater was trained. 
\par Generally, the field of meta-learning or ``learning to learn'' in the context of neural network optimization has been studied and theorized for quite some time \cite{298591} \cite{Hochreiter2001Meta}. For example, \cite{886220} represents a learning update rule as part of the neural network being used for prediction and found that more complicated (depth 20-30) neural network weight-updates performed better for linearly separable learning tasks. On the other hand, genetic algorithms have also been shown to learn effective neural network updaters in the work of \cite{CHALMERS199181}. 


Another approach to meta-learning was implemented in \cite{bello2017neuraloptimizersearchreinforcement}; their framework uses a controller to generate strings (which are specially designed to not contain parentheses) that define the weight update rules. Each update rule applies two separate unary operators on two separate operands, and combines the resulting two operands with a binary operator to form a primitive of the form 
 $\Delta w = \lambda \cdot b( u_1(o_1) , u_2(o_2) )$.
While the domain-specific language (DSL) the authors created is not expansive enough to express all mathematical equations, they show that it is expansive enough to represent competitively-performing optimizers on different machine-learning tasks. 

\par The work of \cite{bello2017neuraloptimizersearchreinforcement} points toward studies most relevant to our work; those which directly investigate discovering \emph{explicit, symbolic weight-update rules} from data. Most closely related to our work, following in the spirit of Auto-ML Zero \cite{real2020automlzeroevolvingmachinelearning}, the authors of \cite{chen2023symbolicdiscoveryoptimizationalgorithms} strongly shift the paradigm from neural network weight-update rules towards the symbolic discovery over the space of \emph{algorithms} used to update the weights of a neural network. Specifically, the authors of \cite{chen2023symbolicdiscoveryoptimizationalgorithms} perform a search over the space of possible \texttt{train} program-functions on a variety of tasks. When considering the performance of their resulting \emph{Lion} (\emph{Evo\textbf{L}ved S\textbf{i}gn M\textbf{o}me\textbf{n}tum}) algorithm, it is important to note that the authors populated the search space with adaptive algorithms such as \texttt{AdamW} and softly constrained the \texttt{train} programs to depend on first and second non-central moments of the gradient-vector at iteration $t$. However, just like in \cite{bello2017neuraloptimizersearchreinforcement}, they enforce the weights to update as:
\begin{equation}
    w_{j, m, t=\tau} = w_{j, m, t=\tau-1} - \hat{f}(\texttt{Operands}),
\end{equation}
where $\hat{f}$ here denotes the symbolic weight-update rule candidate, and the set of \texttt{Operands} is included in table \ref{tab:operands}. 
\par In this work, we only include the operators and operands used to \emph{form} adaptive algorithms instead of explicitly inserting the algorithms themselves into the equation population. Additionally, we frame the problem as a search over weight-update rules of the form\footnote{Note that \texttt{Operands} contains the previous weight $w_{j, m, t=\tau-1}$.}:
\begin{equation}
    w_{j, m, t=\tau} = \hat{f}(\texttt{Operands}).
\end{equation}


In our work, we conduct a search over basis-operands derived from mainstream weight-update rules (table \ref{tab:operands}) to derive a population of explicit, compact, and interpretable weight-update-rule equations. Then, we test their generalizability by seeding initial genetic populations for subsequent benchmarks with weight-update rules found previously.

\section{Experiments}\label{sec:experiments}
In this section, we compare established neural-network update rules against update rules discovered through symbolic regression on a collection of symbolic-regression benchmark problems.

\subsection{Benchmarks}\label{subsec:benchmarks}
As discussed thoroughly in \cite{dahl2023benchmarkingneuralnetworktraining}, there exist numerous ``domain-specific benchmarks'' to quantify neural networks' performance on specific tasks such as image/text classification, prediction, etc. Here, we simply aim to investigate the performance of symbolically generated weight-update rules for learning different symbolic regression benchmark-functions from literature \cite{hemberg2008pre} \cite{udrescu2020ai}.
\par In section \ref{subsec:established_weight_update_grid_search}, we perform a grid search over a range of hyperparameters for each established weight-update rule. Then, in section \ref{subsubsec:symbregressbenchmarks}, we use genetic-programming based symbolic regression for each of the 30 neural net-benchmark combinations\footnote{10 benchmark expressions (see tables \ref{tab:Hemberg2008PreIP_results} and \ref{tab:AI_Feynman_Benchmark_Equations}) $\times$ 3 Neural networks (see figure \ref{fig:feed_forward_neural_net_architectures_considered}) = 30 experiments.} to learn a population of weight-update rules that perform better than the best performing established weight-update rules. In all cases, to make repeated evaluation of thousands of candidate update rules computationally feasible, we restrict all weight-update rules to execute for exactly 10 epochs, and we terminate early if any of the weights become numerically undefined.

\subsection{Established Weight-Update Rules Grid-Search}\label{subsec:established_weight_update_grid_search}
\begin{figure}
    \centering
    \begin{center}
        \underline{\textbf{Neural Network 1}} \\[0.2cm]
        \scalebox{0.5}{
        \begin{tikzpicture}
            \node[ellipse, draw, fill=white] (b1) at (0, 4.5) {Input 1};
            \node[ellipse, draw, fill=white] (b2) at (0, 3.5) {Input 2};
            \node[] at (0, 2.6) {$\vdots$}; 
            \node[ellipse, draw, fill=white] (b3) at (0, 1.5) {Input $N$};
            \node[ellipse, draw, fill=white] (a0) at (3, 4) {Sigmoid};
            \node[ellipse, draw, fill=white] (a1) at (3, 2) {Sigmoid};
            \foreach \i in {1, 2, 3}
            {
                \foreach \j in {0, 1}
                {
                    \draw[thick] (b\i.east) -- (a\j.west);
                }
            }
            \node[ellipse, draw, fill=white] at (0, 4.5) {Input 1};
            \node[ellipse, draw, fill=white] at (0, 3.5) {Input 2};
            \node[ellipse, draw, fill=white] at (0, 1.5) {Input $N$};
            \foreach \x in {0,...,6}
            {
                \node[ellipse, fill=white] (A\x) at (6, \x) {\phantom{Sigmoid}};
                \draw[thick] (a0.east) -- (A\x.west);
                \draw[thick] (a1.east) -- (A\x.west);
            }
            \node[ellipse, draw, fill=white] at (3, 4) {Sigmoid};
            \node[ellipse, draw, fill=white] at (3, 2) {Sigmoid};
            
            \foreach \x/\y in {0.5/0, 1.5/1, 2.5/2, 3.5/3, 4.5/4, 5.5/5}
            {
                \node[ellipse] (B\y) at (9, \x) {\phantom{Sigmoid}};
            }
            
            \foreach \x in {0, 1, ..., 5}
            {
                \foreach \y in {0,...,6}
                {
                    \draw[thick] (B\x.west) -- (A\y.east);
                }
            }

            \foreach \x in {0,...,6}
            {
                \node[ellipse, draw, fill=white] at (6, \x) {Sigmoid};
            }
            \node[ellipse, draw, fill=white] (END) at (12, 3) {Output};

            \foreach \x/\y in {0.5/0, 1.5/1, 2.5/2, 3.5/3, 4.5/4, 5.5/5}
            {
                \node[ellipse, draw, fill=white] (C\y) at (9, \x) {Sigmoid};
                \draw[thick] (C\y.east) -- (END.west);
            }
            
        \end{tikzpicture}} \\[0.5cm]
        \underline{\textbf{Neural Network 2}} \\[0.2cm]
        \scalebox{0.5}{
        \begin{tikzpicture}
            \tikzstyle{neuron}=[ellipse, draw, fill=white]
            
            \node[neuron] (b1) at (0,4) {Input 1};
            \node[neuron] (b2) at (0,3) {Input 2};
            \node at (0,2.6) {$\vdots$};
            \node[neuron] (b3) at (0,1) {Input $N$};
            
            \foreach \y [count=\i from 0] in {0,1,2,3,4,5}
              \node[neuron] (L1\i) at (3,\y) {Sigmoid};
            
            \foreach \y [count=\i from 0] in {-1,0,1,2,3,4,5,6}
              \node[neuron] (L2\i) at (6,\y) {Sigmoid};
            
            \node[neuron] (L30) at (9,2.5) {Sigmoid};
            
            \foreach \y [count=\i from 0] in {0.5,1.5,2.5,3.5,4.5}
              \node[neuron] (L4\i) at (12,\y) {None};
            
            \node[neuron] (OUT) at (15,2.5) {Output};
            
            \foreach \i in {1,2,3}
              \foreach \j in {0,...,5}
                \draw[thick] (b\i.east) -- (L1\j.west);
            
            \foreach \i in {0,...,5}
              \foreach \j in {0,...,7}
                \draw[thick] (L1\i.east) -- (L2\j.west);
            
            \foreach \i in {0,...,7}
              \draw[thick] (L2\i.east) -- (L30.west);
            
            \foreach \j in {0,...,4}
              \draw[thick] (L30.east) -- (L4\j.west);
            
            \foreach \j in {0,...,4}
              \draw[thick] (L4\j.east) -- (OUT.west);
            
        \end{tikzpicture}} \\[0.5cm]
        \underline{\textbf{Neural Network 3}} \\[0.2cm]
        \scalebox{0.5}{
        \begin{tikzpicture}
            \tikzstyle{neuron}=[ellipse, draw, fill=white]
            
            \node[neuron] (b1) at (0,4) {Input 1};
            \node[neuron] (b2) at (0,3) {Input 2};
            \node at (0,2.6) {$\vdots$};
            \node[neuron] (b3) at (0,1) {Input $N$};
            
            \foreach \y [count=\i from 0] in {-2,-1,0,1,2,3,4,5,6,7}
              \node[neuron] (L1\i) at (3,\y) {Sigmoid};
            
            \foreach \y [count=\i from 0] in {-1.5,-0.5,0.5,1.5,2.5,3.5,4.5,5.5,6.5}
              \node[neuron] (L2\i) at (6,\y) {Sigmoid};
            
            \foreach \y [count=\i from 0] in {-1,0,1,2,3,4,5,6}
              \node[neuron] (L3\i) at (9,\y) {Sigmoid};
            
            \foreach \y [count=\i from 0] in {-2,-1,0,1,2,3,4,5,6,7}
              \node[neuron] (L4\i) at (12,\y) {None};
            
            \foreach \y [count=\i from 0] in {-1,0,1,2,3,4,5,6}
              \node[neuron] (L5\i) at (15,\y) {None};
            
            \node[neuron] (OUT) at (18,2.5) {Output};
            
            \foreach \i in {1,2,3}
              \foreach \j in {0,...,9}
                \draw[thick] (b\i.east) -- (L1\j.west);
            
            \foreach \i in {0,...,9}
              \foreach \j in {0,...,8}
                \draw[thick] (L1\i.east) -- (L2\j.west);
            
            \foreach \i in {0,...,8}
              \foreach \j in {0,...,7}
                \draw[thick] (L2\i.east) -- (L3\j.west);
            
            \foreach \i in {0,...,7}
              \foreach \j in {0,...,9}
                \draw[thick] (L3\i.east) -- (L4\j.west);
            
            \foreach \i in {0,...,9}
              \foreach \j in {0,...,7}
                \draw[thick] (L4\i.east) -- (L5\j.west);
            
            \foreach \j in {0,...,7}
              \draw[thick] (L5\j.east) -- (OUT.west);
            
            \end{tikzpicture}
        }
    \end{center}
    \caption{The three feed-forward neural network architectures we consider in the experiments section \ref{sec:experiments}. ``Sigmoid'' denotes a neuron with sigmoid-activation function and ``None'' denotes a neuron with no activation function (linear). The $N$ inputs are determined by the number of inputs in the benchmark symbolic functions that these neural network architectures shall approximate.}
    \label{fig:feed_forward_neural_net_architectures_considered}
\end{figure}
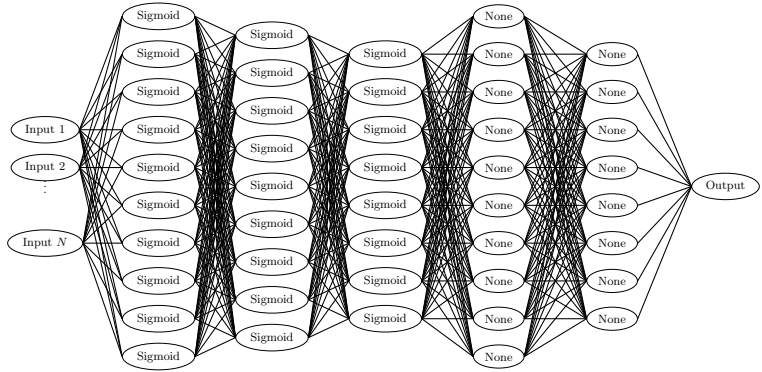

For each of the ten benchmark expressions we consider in this work, denoted in tables \ref{tab:Hemberg2008PreIP_results} and \ref{tab:AI_Feynman_Benchmark_Equations}, we test the performance of 3 simple feed-forward neural network architectures, shown in Figure \ref{fig:feed_forward_neural_net_architectures_considered}, for each weight update rule in table \ref{tab:weight_update_rules_expression_trees}, for the hyperparameter combos denoted in table \ref{tab:hyper_param_grid_searches_done}. 
The results of this hyperparameter sweep are shown in table \ref{tab:grid_sweep_res_est_weight_update_rules}.


\setlength{\tabcolsep}{10pt}
\renewcommand{\arraystretch}{1.3}
\begin{table}[]
    \centering
    \scalebox{0.8}{\begin{tabular}{|c|c|}
    \hline
   \textbf{Weight-Update Rule} & \textbf{Grid-Search Parameters}  \\[0.1cm] \hline
   Gradient Descent      &  $\eta \in \{1.0 \times 10^{-6}, 3.0 \times 10^{-6}, 1.0 \times 10^{-5}, 3.0 \times 10^{-5}, 1.0 \times 10^{-4}, 3.0 \times 10^{-4}, 1.0 \times 10^{-3}, 3.0 \times 10^{-3}\}$ \\ \hline 
   \multirow{ 2}{*}{Heavy-Ball} & $\theta \in \{0.7, 0.8, 0.85, 0.9, 0.95\}$ \\ & $\eta \in \{1.0 \times 10^{-5}, 3.0 \times 10^{-5}, 1.0 \times 10^{-4}, 3.0 \times 10^{-4}, 1.0 \times 10^{-3}\}$ \\ \hline 
   \multirow{ 2}{*}{NAG} & $\theta \in \{0.7, 0.8, 0.85, 0.9, 0.95\}$ \\ & $\eta \in \{1.0 \times 10^{-5}, 3.0 \times 10^{-5}, 1.0 \times 10^{-4}, 3.0 \times 10^{-4}, 1.0 \times 10^{-3}\}$ \\ \hline 
   \multirow{ 2}{*}{Adagrad} & $\epsilon \in \{1.0 \times 10^{-10}, 1.0 \times 10^{-8}, 1.0 \times 10^{-6}, 1.0 \times 10^{-4}\}$ \\ & $\eta \in \{3.0 \times 10^{-4}, 1.0 \times 10^{-3}, 3.0 \times 10^{-3}, 1.0 \times 10^{-2}, 3.0 \times 10^{-2}\}$ \\ \hline
   \multirow{ 3}{*}{RMSProp} & $\epsilon \in \{1.0 \times 10^{-10}, 1.0 \times 10^{-8}, 1.0 \times 10^{-6}\}$ \\ & $\eta \in \{1.0 \times 10^{-6}, 3.0 \times 10^{-6}, 1.0 \times 10^{-5}, 3.0 \times 10^{-5}, 1.0 \times 10^{-4}, 3.0 \times 10^{-4}, 1.0 \times 10^{-3}\}$ \\ & $\gamma \in \{ 0.9, 0.95, 0.99\}$\\ \hline 
   \multirow{ 2}{*}{AdaDelta} & $\epsilon \in \{1.0 \times 10^{-10}, 1.0 \times 10^{-8}, 1.0 \times 10^{-6}, 1.0 \times 10^{-4}\}$ \\ & $\gamma \in \{0.9, 0.95, 0.99, 0.995\}$ \\ \hline
    \multirow{ 4}{*}{Adam} & $\eta \in \{1.0 \times 10^{-6}, 3.0 \times 10^{-6}, 1.0 \times 10^{-5}, 3.0 \times 10^{-5}, 1.0 \times 10^{-4}\}$ \\ & $\epsilon \in \{1.0 \times 10^{-10}, 1.0 \times 10^{-8}, 1.0 \times 10^{-6}, 1.0 \times 10^{-4}\}$ \\ & $\beta_1 \in \{0.8, 0.9, 0.95\}$ \\ & $\beta_2 \in \{0.99, 0.995, 0.999\}$ \\ \hline
    \multirow{ 5}{*}{AdamW} & $\lambda \in \{ 1.0 \times 10^{-6}, 1.0 \times 10^{-5}, 1.0 \times 10^{-4}, 1.0 \times 10^{-3}\}$ \\ & $\eta \in \{1.0 \times 10^{-6}, 3.0 \times 10^{-6}, 1.0 \times 10^{-5}, 3.0 \times 10^{-5}, 1.0 \times 10^{-4}\}$ \\ & $\epsilon \in \{1.0 \times 10^{-10}, 1.0 \times 10^{-8}, 1.0 \times 10^{-6}, 1.0 \times 10^{-4}\}$ \\ & $\beta_1 \in \{0.8, 0.9, 0.95\}$ \\ & $\beta_2 \in \{0.99, 0.995, 0.999\}$ \\ \hline
    \end{tabular}}
    \caption{Hyperparameter grid searches for each weight-update rule considered for each benchmark neural network combo.}
    \label{tab:hyper_param_grid_searches_done}
\end{table}

\setlength{\tabcolsep}{10pt}
\renewcommand{\arraystretch}{1.3}
\begin{table}[]
    \centering
\scalebox{0.64}{\begin{tabular}{|c|c|c|l|l|l|l|l|l|l|l|}
\hline 
\textbf{Benchmark} & \textbf{NN} & \textbf{Best Rule} & $\bm{\eta}$ & $\bm{\theta}$ & $\bm{\gamma}$ & $\bm{\epsilon}$ & $\bm{\beta_1}$ & $\bm{\beta_2}$ & $\bm{\lambda}$ & \textbf{MSE} 
 \\[0.14cm] \hline
Feynman 1 & 1 & heavy ball & $1.00 \times 10^{-3}$ & $9.00 \times 10^{-1}$ & N/A & N/A & N/A & N/A & N/A & $3.51 \times 10^{0}$ \\
Feynman 1 & 2 & heavy ball & $1.00 \times 10^{-3}$ & $8.00 \times 10^{-1}$ & N/A & N/A & N/A & N/A & N/A & $3.54 \times 10^{0}$ \\
Feynman 1 & 3 & AdaDelta & N/A & N/A & $9.95 \times 10^{-1}$ & $1.00 \times 10^{-4}$ & N/A & N/A & N/A & $4.03 \times 10^{0}$ \\
Feynman 2 & 1 & AdaDelta & N/A & N/A & $9.50 \times 10^{-1}$ & $1.00 \times 10^{-4}$ & N/A & N/A & N/A & $1.25 \times 10^{1}$ \\
Feynman 2 & 2 & AdaDelta & N/A & N/A & $9.90 \times 10^{-1}$ & $1.00 \times 10^{-4}$ & N/A & N/A & N/A & $1.49 \times 10^{1}$ \\
Feynman 2 & 3 & AdaDelta & N/A & N/A & $9.95 \times 10^{-1}$ & $1.00 \times 10^{-6}$ & N/A & N/A & N/A & $4.83 \times 10^{0}$ \\
Feynman 3 & 1 & heavy ball & $3.00 \times 10^{-4}$ & $9.50 \times 10^{-1}$ & N/A & N/A & N/A & N/A & N/A & $2.89 \times 10^{8}$ \\
Feynman 3 & 2 & AdaGrad & $3.00 \times 10^{-3}$ & N/A & N/A & N/A & N/A & N/A & N/A & $2.93 \times 10^{8}$ \\
Feynman 3 & 3 & AdaGrad & $1.00 \times 10^{-3}$ & N/A & N/A & N/A & N/A & N/A & N/A & $2.92 \times 10^{8}$ \\
Feynman 4 & 1 & heavy ball & $3.00 \times 10^{-4}$ & $9.00 \times 10^{-1}$ & N/A & N/A & N/A & N/A & N/A & $5.41 \times 10^{0}$ \\
Feynman 4 & 2 & basic & $1.00 \times 10^{-3}$ & N/A & N/A & N/A & N/A & N/A & N/A & $7.77 \times 10^{0}$ \\
Feynman 4 & 3 & AdaDelta & N/A & N/A & $9.00 \times 10^{-1}$ & $1.00 \times 10^{-4}$ & N/A & N/A & N/A & $2.09 \times 10^{0}$ \\
Feynman 5 & 1 & AdaDelta & N/A & N/A & $9.50 \times 10^{-1}$ & $1.00 \times 10^{-4}$ & N/A & N/A & N/A & $1.82 \times 10^{1}$ \\
Feynman 5 & 2 & AdaDelta & N/A & N/A & $9.00 \times 10^{-1}$ & $1.00 \times 10^{-4}$ & N/A & N/A & N/A & $2.07 \times 10^{1}$ \\
Feynman 5 & 3 & AdaDelta & N/A & N/A & $9.00 \times 10^{-1}$ & $1.00 \times 10^{-4}$ & N/A & N/A & N/A & $1.65 \times 10^{1}$ \\
Hemberg  1 & 1 & NAG & $1.00 \times 10^{-3}$ & $9.00 \times 10^{-1}$ & N/A & N/A & N/A & N/A & N/A & $3.17 \times 10^{-1}$ \\
Hemberg  1 & 2 & NAG & $1.00 \times 10^{-3}$ & $7.00 \times 10^{-1}$ & N/A & N/A & N/A & N/A & N/A & $3.89 \times 10^{-1}$ \\
Hemberg  1 & 3 & NAG & $1.00 \times 10^{-4}$ & $8.00 \times 10^{-1}$ & N/A & N/A & N/A & N/A & N/A & $3.79 \times 10^{-1}$ \\
Hemberg  2 & 1 & NAG & $3.00 \times 10^{-4}$ & $9.50 \times 10^{-1}$ & N/A & N/A & N/A & N/A & N/A & $5.47 \times 10^{2}$ \\
Hemberg  2 & 2 & heavy ball & $1.00 \times 10^{-3}$ & $8.00 \times 10^{-1}$ & N/A & N/A & N/A & N/A & N/A & $4.91 \times 10^{2}$ \\
Hemberg  2 & 3 & NAG & $1.00 \times 10^{-4}$ & $7.00 \times 10^{-1}$ & N/A & N/A & N/A & N/A & N/A & $4.09 \times 10^{2}$ \\
Hemberg  3 & 1 & AdaDelta & N/A & N/A & $9.90 \times 10^{-1}$ & $1.00 \times 10^{-4}$ & N/A & N/A & N/A & $1.49 \times 10^{1}$ \\
Hemberg  3 & 2 & heavy ball & $1.00 \times 10^{-3}$ & $8.00 \times 10^{-1}$ & N/A & N/A & N/A & N/A & N/A & $9.05 \times 10^{0}$ \\
Hemberg  3 & 3 & AdaDelta & N/A & N/A & $9.90 \times 10^{-1}$ & $1.00 \times 10^{-4}$ & N/A & N/A & N/A & $6.97 \times 10^{0}$ \\
Hemberg  4 & 1 & NAG & $1.00 \times 10^{-3}$ & $8.50 \times 10^{-1}$ & N/A & N/A & N/A & N/A & N/A & $5.56 \times 10^{3}$ \\
Hemberg  4 & 2 & heavy ball & $1.00 \times 10^{-4}$ & $8.50 \times 10^{-1}$ & N/A & N/A & N/A & N/A & N/A & $6.01 \times 10^{3}$ \\
Hemberg  4 & 3 & AdaDelta & N/A & N/A & $9.50 \times 10^{-1}$ & $1.00 \times 10^{-4}$ & N/A & N/A & N/A & $3.34 \times 10^{3}$ \\
Hemberg  5 & 1 & heavy ball & $3.00 \times 10^{-4}$ & $9.50 \times 10^{-1}$ & N/A & N/A & N/A & N/A & N/A & $5.67 \times 10^{3}$ \\
Hemberg  5 & 2 & heavy ball & $1.00 \times 10^{-4}$ & $8.50 \times 10^{-1}$ & N/A & N/A & N/A & N/A & N/A & $5.86 \times 10^{3}$ \\
Hemberg  5 & 3 & AdaDelta & N/A & N/A & $9.50 \times 10^{-1}$ & $1.00 \times 10^{-4}$ & N/A & N/A & N/A & $3.77 \times 10^{3}$ \\
\hline
\end{tabular}}
    \caption{Results of the hyperparameter grid sweep conducted; see table \ref{tab:hyper_param_grid_searches_done}. Benchmarks Feynman 1-5 denote the corresponding equations 1-5 in table \ref{tab:AI_Feynman_Benchmark_Equations}, and benchmarks Hemberg 1-5 denote the corresponding equations 1-5 in table \ref{tab:Hemberg2008PreIP_results}.}
    \label{tab:grid_sweep_res_est_weight_update_rules}
\end{table}


\subsection{Symbolic Regression Benchmarks}\label{subsubsec:symbregressbenchmarks}

Following \cite{finkelstein2024generalizedfixeddepthprefixpostfix}, we apply symbolic regression to discover neural network update rules on benchmark expressions from \cite{hemberg2008pre} and \cite{udrescu2020ai} with varying complexity/depth of expression trees. The ground-truth equations, 10 in total, are shown in tables \ref{tab:Hemberg2008PreIP_results} and \ref{tab:AI_Feynman_Benchmark_Equations}.
\begin{table} 
    \centering
    \scalebox{0.8}{
    \begin{tabular}{|l|l|l|l|l|}
\hline 
\# & Expression & Expression Tree & Depth $N$ & \# of Inputs \\ \hline 
 1 &   $8/(2+x^2+y^2)$ & \href{https://github.com/edfink234/Alpha-Zero-Symbolic-Regression/blob/13f3cc08ec72008eb735a00c14084f9b0af08293/Hemberg2008Expressions/expression_tree_Hemberg2008_expr_1.pdf}{Hemberg 1} & 4 & 2 \\[0.2cm]
 2 &    $x^3\cdot(x-1) + y\cdot(y/2-1)$ & \href{https://github.com/edfink234/Alpha-Zero-Symbolic-Regression/blob/13f3cc08ec72008eb735a00c14084f9b0af08293/Hemberg2008Expressions/expression_tree_Hemberg2008_expr_2.pdf}{Hemberg 2} & 4 & 2 \\[0.2cm]
 3 & $x^3/5 + y^3/2 - y - x$ & \href{https://github.com/edfink234/Alpha-Zero-Symbolic-Regression/blob/13f3cc08ec72008eb735a00c14084f9b0af08293/Hemberg2008Expressions/expression_tree_Hemberg2008_expr_3.pdf}{Hemberg 3} & 5 & 2 \\[0.2cm]
 4 & $\frac{30\cdot x^2}{(10-x)\cdot y^2}+x^4 - x^3 + \frac{y^2}{2} - y + \frac{8}{2+x^2+y^2} + x$ & \href{https://github.com/edfink234/Alpha-Zero-Symbolic-Regression/blob/13f3cc08ec72008eb735a00c14084f9b0af08293/Hemberg2008Expressions/expression_tree_Hemberg2008_expr_4.pdf}{Hemberg 4} & 9 & 2 \\[0.2cm]
 5 & $\frac{30\cdot x^2}{(10-x)\cdot y^2}+x^4 - \frac{4}{5}x^3 + \frac{y^2}{2} - 2y + \frac{8}{2+x^2+y^2} + \frac{y^3}{2} - x$ & \href{https://github.com/edfink234/Alpha-Zero-Symbolic-Regression/blob/13f3cc08ec72008eb735a00c14084f9b0af08293/Hemberg2008Expressions/expression_tree_Hemberg2008_expr_5.pdf}{Hemberg 5} & 10 & 2 \\[0.2cm] \hline
\end{tabular}}
    \caption{Expressions from \cite{hemberg2008pre} considered in section \ref{subsec:benchmarks}.}
    \label{tab:Hemberg2008PreIP_results}
\end{table}

\begin{table} 
    \centering
    \scalebox{0.82}{
    \begin{tabular}{|l|l|l|l|l|}
\hline 
\# & Expression & Expression Tree & Depth $N$ & \# of Inputs \\ \hline 
 1 & $x = \frac{q\cdot E_f}{m\cdot \left(\omega_0^2 - \omega^2\right)}$ & \href{https://edfink234.github.io/AIFeynmanExpressionTrees/AIFeynmanExpressionTrees/AIFeynmanExpressionTrees\#figcaption61}{Feynman 1} & 4 & 5 \\[0.2cm]
 2 & $F = \frac{G\cdot m_1 \cdot m_2}{\left(x_2 - x_1\right)^2 + \left(y_2 - y_1\right)^2 +\left(z_2 - z_1\right)^2}$ & \href{https://edfink234.github.io/AIFeynmanExpressionTrees/AIFeynmanExpressionTrees/AIFeynmanExpressionTrees\#figcaption5}{Feynman 2} & 5 & 9 \\[0.2cm]
 3 & $A = \left(\frac{Z_1 \cdot Z_2 \cdot \alpha \cdot \hbar \cdot c}{4\cdot E_n \cdot \sin^2\left(\theta/2\right)}\right)^2$ & \href{https://edfink234.github.io/AIFeynmanExpressionTrees/AIFeynmanExpressionTrees/AIFeynmanExpressionTrees\#figcaption101}{Feynman 3} & 6 & 7 \\[0.2cm]
 4 & $f = \frac{\mu_m\cdot H}{k_b \cdot T} + \frac{\mu_m\cdot \alpha}{\epsilon \cdot c^2 \cdot k_b \cdot T} \cdot M$ & \href{https://edfink234.github.io/AIFeynmanExpressionTrees/AIFeynmanExpressionTrees/AIFeynmanExpressionTrees\#figcaption82}{Feynman 4} & 7 & 8 \\[0.2cm]
 5 & $k = \frac{m \cdot k_G}{L^2}\cdot \left(1+\sqrt{\left(1+\frac{2 \cdot E_n \cdot L^2}{m\cdot k_G^2}\right)}\cdot \cos\left(\theta_1-\theta_2\right)\right)$ & \href{https://edfink234.github.io/AIFeynmanExpressionTrees/AIFeynmanExpressionTrees/AIFeynmanExpressionTrees\#figcaption102}{Feynman 5} & 8 & 6 \\[0.2cm] \hline
\end{tabular}}
    \caption{Expressions from \cite{udrescu2020ai} considered in section \ref{subsec:benchmarks}.}
    \label{tab:AI_Feynman_Benchmark_Equations}
\end{table}
For each of the 30 benchmark/neural network combinations\footnote{denoted by the first 2 columns of Table \ref{tab:grid_sweep_res_est_weight_update_rules}}, we allow a maximum genetic population size of 100 candidate weight-update rules split across all threads such that each thread performs an independent search, as preliminary experiments indicated that larger populations produced similar solutions at substantially greater computational cost. The tree-depth for each generated weight update rule is fixed to 5; thus, there are $\approx 1.71\times 10^{65}$ weight-update rules that can possibly be generated in our study\footnote{The number of unique equations for a given depth is computed as $N_{\mathrm{depth-}i} = N_U \cdot N_{\mathrm{depth-}(i-1)} + N_B  \left( N_{\mathrm{depth}-(i-1)}^2 + 2 \cdot N_{\mathrm{depth-}(i-1)}\left(\sum_{j = 0}^{i-2}N_{\mathrm{depth-}j}\right)\right)$, where $N_{\mathrm{depth-1}} = N_U\cdot N_L + N_B \cdot N_L^2$ and $N_{\mathrm{depth-0}} = \text{ }N_L$, with $N_B = 5$, $N_L = 22$, and $N_U = 9$. See a live code \href{https://www.sololearn.com/en/compiler-playground/cI8qd1dMjMog}{here}.}. For each experiment, we first attempt to seed the starting populations with the best weight-update rules found in the previous experiment; however, if evolving this population fails to produce competitive update rules for a given benchmark, we fall back to evolving a fresh random population. For each experiment, evolution continued until improvements in the best fitness became negligible over an extended period. Runs were also terminated early if 100 expressions were discovered that outperformed the corresponding best hyperparameter-tuned baseline optimizer.

\section{Results}\label{sec:Results}

Table \ref{tab:results} shows the results of our 30 experiments. Symbolic regression discovered update rules that outperformed the best hyperparameter-tuned baseline in 25 of 30 benchmark/neural network combinations, yielding an aggregate MSE reduction of 44.47\% across the improved cases. The five failure cases were concentrated in the largest architecture (Neural Network 3), suggesting that the discovered rules may become less competitive as model complexity increases.

Although the best discovered update equations (available \href{https://cb1658.github.io/BenchmarkRules/}{here}) do not appear to share a single obvious symbolic structure, several qualitative trends emerge. Many of the discovered rules incorporate nonlinearities such as trigonometric functions, hyperbolic tangent functions, exponentials, and rational expressions involving moving-average statistics. Several also resemble familiar optimizer design motifs, including momentum-like terms, adaptive normalization, and variance scaling, despite these structures not being explicitly enforced during search.

\par At the same time, the discovered expressions are not entirely unrelated. Among the best expressions found by symbolic regression for each benchmark, four canonical expression signatures reappear across multiple benchmarks, see Table \ref{tab:repeat_expressions}, all of which incorporate \emph{accumulated} gradient statistics over raw gradients.

\begin{table}[]
    \centering
    \begin{tabular}{ll}
    \toprule
    Benchmarks & Signature \\
    \midrule
    B2, B3 & $C \mu_{j,m,t=\tau} + C$ \\
    B3, B4 & $C \Delta w_{j, m, t=\tau} + C$ \\
    B26, B28, B29 & $C t \left(C \mu_{j,m,t=\tau} + C\right)^{C}$ \\
    B28, B29 & $C t \left(C \mu_{j,m,t=\tau} + C y_j + C\right)^{C}$ \\
    \bottomrule
    \end{tabular}
    \caption{Unique weight-update expression signatures that reappear as part of the best expressions found by symbolic regression across benchmarks, where $C$ represents a numerical-value and the operands are denoted in table \ref{tab:operands}.}
    \label{tab:repeat_expressions}
\end{table}

Many algebraically distinct expressions nevertheless achieved comparable predictive performance, suggesting that the search space contains large equivalence classes of effective update dynamics. This observation may partially explain why symbolic regression was able to repeatedly discover update equations that outperform standard hand-designed optimizers despite the overall structural diversity of the final expressions.

Another notable observation is that the symbolic regression-discovered rules often remained relatively compact despite the large search space. This suggests that useful optimization dynamics may frequently admit low-complexity symbolic representations, even when their functional forms differ substantially across tasks.
\setlength{\tabcolsep}{10pt}
\renewcommand{\arraystretch}{1.3}
\begin{table}[]
    \centering
    \scalebox{0.6}{\begin{tabular}{|c|c|c|c|c|c|c|c|}
\hline
\textbf{ID} & \textbf{Benchmark} & \textbf{NN} &
\textbf{Best Est. Rule} &
\textbf{Best Est. Rule MSE} &
\textbf{Best Found Rule} &
\textbf{Best Found Rule MSE} &
\textbf{\# Better Exprs}
\\[0.14cm] \hline
1 & Feynman 1 & 1 & heavy ball & $3.51 \times 10^{0}$ & \href{https://cb1658.github.io/BenchmarkRules/\#benchmark-1}{Benchmark 1} & $2.46 \times 10^{0}$ & {100} \\
2 & Feynman 1 & 2 & heavy ball & $3.54 \times 10^{0}$ & \href{https://cb1658.github.io/BenchmarkRules/\#benchmark-2}{Benchmark 2} & $2.53 \times 10^{0}$ & {100} \\
3 & Feynman 1 & 3 & AdaDelta & $4.03 \times 10^{0}$ & \href{https://cb1658.github.io/BenchmarkRules/\#benchmark-3}{Benchmark 3} & $2.54 \times 10^{0}$ & {100} \\
4 & Feynman 2 & 1 & AdaDelta & $1.25 \times 10^{1}$ & \href{https://cb1658.github.io/BenchmarkRules/\#benchmark-4}{Benchmark 4} & $7.75 \times 10^{0}$ & {98} \\
5 & Feynman 2 & 2 & AdaDelta & $1.49 \times 10^{1}$ & \href{https://cb1658.github.io/BenchmarkRules/\#benchmark-5}{Benchmark 5} & $9.32 \times 10^{0}$ & {100} \\
6 & Feynman 2 & 3 & AdaDelta & $4.83 \times 10^{0}$ & \href{https://cb1658.github.io/BenchmarkRules/\#benchmark-6}{Benchmark 6} & $8.00 \times 10^{0}$ & {0} \\
7 & Feynman 3 & 1 & heavy ball & $2.89 \times 10^{8}$ & \href{https://cb1658.github.io/BenchmarkRules/\#benchmark-7}{Benchmark 7} & $2.37 \times 10^{8}$ & {100} \\
8 & Feynman 3 & 2 & AdaGrad & $2.93 \times 10^{8}$ & \href{https://cb1658.github.io/BenchmarkRules/\#benchmark-8}{Benchmark 8} & $2.88 \times 10^{6}$ & {100} \\
9 & Feynman 3 & 3 & AdaGrad & $2.92 \times 10^{8}$ & \href{https://cb1658.github.io/BenchmarkRules/\#benchmark-9}{Benchmark 9} & $2.45 \times 10^{8}$ & {100} \\
10 & Feynman 4 & 1 & heavy ball & $5.41 \times 10^{0}$ & \href{https://cb1658.github.io/BenchmarkRules/\#benchmark-10}{Benchmark 10} & $2.78 \times 10^{0}$ & {100} \\
11 & Feynman 4 & 2 & basic & $7.77 \times 10^{0}$ & \href{https://cb1658.github.io/BenchmarkRules/\#benchmark-11}{Benchmark 11} & $3.63 \times 10^{0}$ & {98} \\
12 & Feynman 4 & 3 & AdaDelta & $2.09 \times 10^{0}$ & \href{https://cb1658.github.io/BenchmarkRules/\#benchmark-12}{Benchmark 12} & $3.29 \times 10^{0}$ & {0} \\
13 & Feynman 5 & 1 & AdaDelta & $1.82 \times 10^{1}$ & \href{https://cb1658.github.io/BenchmarkRules/\#benchmark-13}{Benchmark 13} & $1.37 \times 10^{1}$ & {35} \\
14 & Feynman 5 & 2 & AdaDelta & $2.07 \times 10^{1}$ & \href{https://cb1658.github.io/BenchmarkRules/\#benchmark-14}{Benchmark 14} & $1.54 \times 10^{1}$ & {100} \\
15 & Feynman 5 & 3 & AdaDelta & $1.65 \times 10^{1}$ & \href{https://cb1658.github.io/BenchmarkRules/\#benchmark-15}{Benchmark 15} & $1.41 \times 10^{1}$ & {7} \\
16 & Hemberg 1 & 1 & NAG & $3.17 \times 10^{-1}$ & \href{https://cb1658.github.io/BenchmarkRules/\#benchmark-16}{Benchmark 16} & $2.73 \times 10^{-1}$ & {48} \\
17 & Hemberg 1 & 2 & NAG & $3.89 \times 10^{-1}$ & \href{https://cb1658.github.io/BenchmarkRules/\#benchmark-17}{Benchmark 17} & $2.84 \times 10^{-1}$ & {40} \\
18 & Hemberg 1 & 3 & NAG & $3.79 \times 10^{-1}$ & \href{https://cb1658.github.io/BenchmarkRules/\#benchmark-18}{Benchmark 18} & $2.69 \times 10^{-1}$ & {17} \\
19 & Hemberg 2 & 1 & NAG & $5.47 \times 10^{2}$ & \href{https://cb1658.github.io/BenchmarkRules/\#benchmark-19}{Benchmark 19} & $3.80 \times 10^{2}$ & {100} \\
20 & Hemberg 2 & 2 & heavy ball & $4.91 \times 10^{2}$ & \href{https://cb1658.github.io/BenchmarkRules/\#benchmark-20}{Benchmark 20} & $3.18 \times 10^{2}$ & {19} \\
21 & Hemberg 2 & 3 & NAG & $4.09 \times 10^{2}$ & \href{https://cb1658.github.io/BenchmarkRules/\#benchmark-21}{Benchmark 21} & $3.73 \times 10^{2}$ & {4} \\
22 & Hemberg 3 & 1 & AdaDelta & $1.49 \times 10^{1}$ & \href{https://cb1658.github.io/BenchmarkRules/\#benchmark-22}{Benchmark 22} & $1.12 \times 10^{1}$ & {25} \\
23 & Hemberg 3 & 2 & heavy ball & $9.05 \times 10^{0}$ & \href{https://cb1658.github.io/BenchmarkRules/\#benchmark-23}{Benchmark 23} & $1.60 \times 10^{1}$ & {0} \\
24 & Hemberg 3 & 3 & AdaDelta & $6.97 \times 10^{0}$ & \href{https://cb1658.github.io/BenchmarkRules/\#benchmark-24}{Benchmark 24} & $1.68 \times 10^{1}$ & {0} \\
25 & Hemberg 4 & 1 & NAG & $5.56 \times 10^{3}$ & \href{https://cb1658.github.io/BenchmarkRules/\#benchmark-25}{Benchmark 25} & $3.41 \times 10^{3}$ & {100} \\
26 & Hemberg 4 & 2 & heavy ball & $6.01 \times 10^{3}$ & \href{https://cb1658.github.io/BenchmarkRules/\#benchmark-26}{Benchmark 26} & $2.94 \times 10^{3}$ & {100} \\
27 & Hemberg 4 & 3 & AdaDelta & $3.34 \times 10^{3}$ & \href{https://cb1658.github.io/BenchmarkRules/\#benchmark-27}{Benchmark 27} & $3.51 \times 10^{3}$ & {0} \\
28 & Hemberg 5 & 1 & heavy ball & $5.67 \times 10^{3}$ & \href{https://cb1658.github.io/BenchmarkRules/\#benchmark-28}{Benchmark 28} & $2.82 \times 10^{3}$ & {100} \\
29 & Hemberg 5 & 2 & heavy ball & $5.86 \times 10^{3}$ & \href{https://cb1658.github.io/BenchmarkRules/\#benchmark-29}{Benchmark 29} & $3.11 \times 10^{3}$ & {100} \\
30 & Hemberg 5 & 3 & AdaDelta & $3.77 \times 10^{3}$ & \href{https://cb1658.github.io/BenchmarkRules/\#benchmark-30}{Benchmark 30} & $3.24 \times 10^{3}$ & {3} \\
\hline
    \end{tabular}}
    \caption{Results of running the symbolic regression searches delineated in sub-section \ref{subsubsec:symbregressbenchmarks}. For 25/30 experiments, symbolic regression was able to find a weight-update rule that performed better than the best hyperparameter-tuned baseline rule in sub-section \ref{subsec:established_weight_update_grid_search}. For the 5 experiments where symbolic regression failed to find a better rule, 4 of them utilized neural network architecture 3 from Figure \ref{fig:feed_forward_neural_net_architectures_considered} and the remaining failure case was architecture 2 on Hemberg 3.}
    \label{tab:results}
\end{table}

\section{Limitations}

The experiments in this paper should be interpreted as a preliminary study rather than a complete optimizer benchmark. The neural networks considered are small, training is restricted to 10 epochs, and performance is evaluated on symbolic regression benchmark functions rather than large-scale machine-learning datasets. Moreover, the symbolic regression search is stochastic, and the discovered expressions are selected based on empirical performance rather than theoretical convergence guarantees. Thus, while the results demonstrate that symbolic regression can discover effective, compact update rules in this setting, further experiments are needed to determine how these rules generalize to larger architectures, longer training horizons, classification tasks, and standard deep-learning benchmarks.

\section{Conclusion and Outlook}

We presented a symbolic regression framework for searching the space of explicit feed-forward neural network weight-update rules. Using a fixed set of operators and optimizer-derived operands, the method discovered update rules that outperformed the best hyperparameter-tuned established optimizer in 25 of 30 benchmark/neural network combinations. Across the improved cases, the aggregate reduction in MSE was 44.47\%.

The discovered rules were not structurally uniform, despite our strategy of seeding genetic populations with rules discovered in previous benchmarks, suggesting that the space of effective update rules may be heterogeneous and problem-dependent. Nevertheless, many rules combined recognizable optimizer components, including adaptive moment quantities, gradient-normalization terms, momentum-like variables, and nonlinear transformations. This indicates that symbolic regression may be useful not only for optimizer discovery, but also for generating interpretable candidate update rules that can be inspected and simplified. 

Future work should evaluate these rules on larger neural networks, longer training runs, classification datasets, and repeated random seeds. It would also be useful to add complexity penalties, stability constraints, and post-hoc algebraic simplification to the search procedure, so that discovered optimizers are both performant and easier to analyze.

\bibliographystyle{splncs04}
\bibliography{PrefixPostfixPaper}


\end{document}